\documentclass[10pt,twocolumn,letterpaper]{article}

\usepackage{cvpr}
\usepackage{times}
\usepackage{epsfig}
\usepackage{graphicx}
\usepackage{amsmath}
\usepackage{amssymb}
\usepackage{subfigure}
\usepackage{epstopdf}
\usepackage{bbm}

\usepackage[breaklinks=true,bookmarks=false]{hyperref}

\cvprfinalcopy 

\def\cvprPaperID{2368} 

\ifcvprfinal\fi
\begin{document}

\title{Real-time Action Recognition with Enhanced Motion Vector CNNs}

\author{Bowen Zhang$^{1,2}$ \quad \quad Limin Wang$^{1,3}$ \quad \quad Zhe Wang$^{1}$ \quad \quad Yu Qiao$^{1}\thanks{Corresponding author.}$ \quad \quad Hanli Wang$^2$ \\
\small $^{1}$Shenzhen key lab of Comp. Vis. \& Pat. Rec.,  Shenzhen Institutes of Advanced Technology, CAS, China \\
\small $^{2}$Key Laboratory of Embedded System and Service Computing, Ministry of Education, Tongji University, Shanghai, China\\
\small $^{3}$Computer Vision Lab, ETH Zurich, Switzerland
}

\maketitle
\thispagestyle{empty}

\begin{abstract}
  The deep two-stream architecture \cite{simonyan2014two} exhibited excellent performance on video based action recognition. The most computationally expensive step in this approach comes from the calculation of optical flow which prevents it to be real-time. This paper accelerates this architecture by replacing optical flow with motion vector which can be obtained directly from compressed videos without extra calculation. However, motion vector lacks fine structures, and contains noisy and inaccurate motion patterns, leading to the evident degradation of recognition performance. Our key insight for relieving this problem is that optical flow and motion vector are inherent correlated. Transferring the knowledge learned with optical flow CNN to motion vector CNN can significantly boost the performance of the latter. Specifically, we introduce three strategies for this, initialization transfer, supervision transfer and their combination. Experimental results show that our method achieves comparable recognition performance to the state-of-the-art, while our method can process 390.7 frames per second, which is 27 times faster than the original two-stream method.

\end{abstract}

\section{Introduction}

Action recognition aims to enable computer automatically recognize human action in real world video. Recent years have witnessed extensive research efforts and  significant research progresses in this area.
The existence of large action datasets and worldwide competitions, like UCF101 \cite{soomro2012ucf101}, HMDB51 \cite{kuehne2011hmdb}, and THUMOS14 \cite{THUMOS14} promote researches in this area. Early approaches in this area utilize a Bag-of-Visual-Words paradigm and its variants \cite{DBLP:journals/corr/PengWWQ14}, which mainly consists of feature extraction, feature encoding, and classification steps. The performance of these approaches highly depends on the hand crafted features. Previous studies show that iDT descriptors and Fisher vector representation yield superior performance on various datasets \cite{wang2013action}. More recently researchers exploit deep learning for action recognition. One successful example along this line is the two-stream framework \cite{simonyan2014two} which utilizes both RGB CNN and optical flow CNN for classification and achieves the state-of-the-art performance on several large action datasets. However, two-stream CNNs cannot process videos at real-time. 
The calculation of optical flow is time consuming which hinders the processing speed of two-stream CNNs.
\begin{figure}
\begin{center}
\includegraphics[scale=.6]{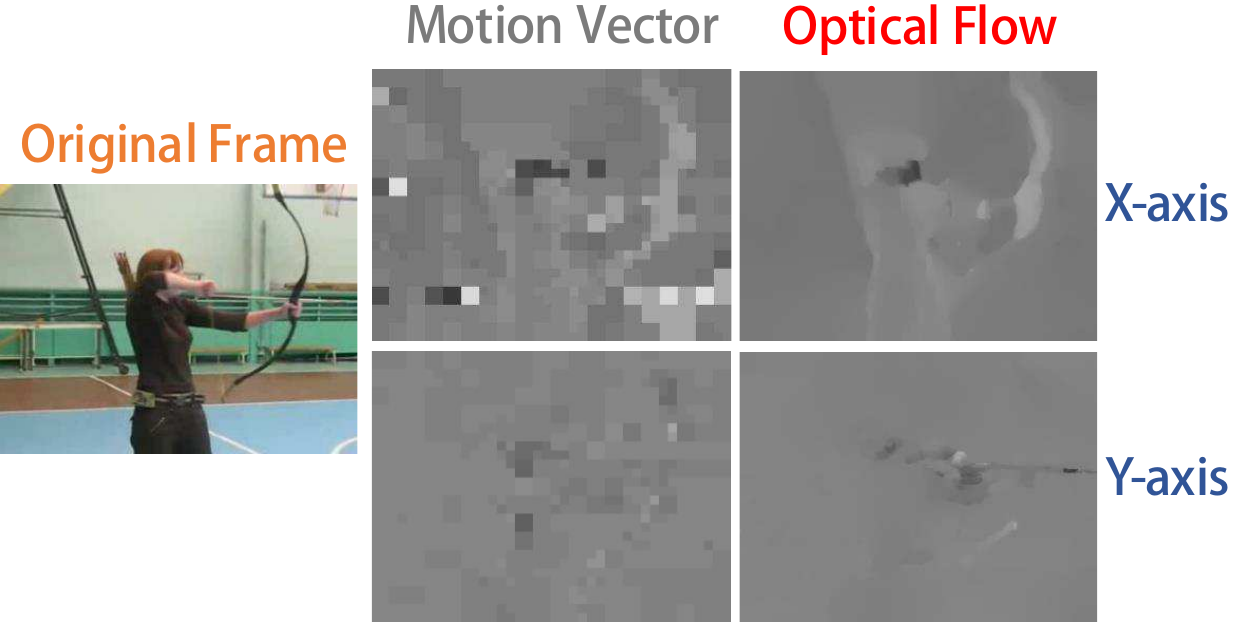}
\end{center}
\vspace {-2mm}
   \caption{Comparison of motion vector and optical flow in x and y components. We can see that motion vector contains lots of noisy movement information and it is much coarser than optical flow.}
\label{fig:Comparison of Optical Flow and MVS}
\vspace {-5mm}
\end{figure}

This paper aims to develop a real-time action recognition method with high performance based on the successful two-stream framework. This is challenging, since optical flow itself is computationally expensive and cannot be estimated in real-time with most current algorithms \cite{brox2011large,farneback2003two}. It takes 60 ms \cite{brox2011large} to calculate optical flows per frame in K40 GPU, which is far from the requirement of real-time processing. To circumvent this difficulty, instead of using optical flow, this paper leverages motion vector as the input of CNN, which can be decoded directly from standard video compressed files with very low computational cost.

Motion vectors represent movement patterns of image blocks which resemble optical flows in terms of describing local motions. Early research \cite{kantorov2014efficient} indicates that motion vectors include useful information for action recognition. However, the purpose of motion vector is not to unveil the temporal relationship of two macro blocks as accurate as possible, but to exploit temporal redundancy between adjacent frames to reduce the bit rate in video compression. Thus motion vector only contains coarse movement patterns which are usually not precise. Moreover, motion vector lacks fine motion information of pixels. Directly replacing optical flows with motion vectors will severely degrade recognition performance of CNN as observed in our experiments.

Here our key insight to improve the recognition performance of motion vector CNN is that optical flow and motion vector share some similar characteristics which allows us to transfer the fine features/knowledge learned in optical flow CNN  (OF-CNN) to that of motion vectors (MV-CNN). Both optical flow and motion vector can be extracted at each frame, and both of them contain motion information of local regions. Motion vector contains coarse and inaccurate motion information while optical flow carries fine and accurate ones. Due to the fine quality of optical flow fields, OF-CNN can learn elaborate filters and achieve better performance than MV-CNN. These facts inspire us to leverage the knowledge learned with OF-CNN to enhance MV-CNN. More specifically, we take OF-CNN learned in optical flow domain as a teacher net, and teach student MV-CNN in spirit of the knowledge distillation techniques proposed by \cite{hinton2015distilling}. We call this new CNN as optical flow enhanced motion vector CNN. Note we only require optical flow in the training phase, while the testing can be conducted with motion vector solely. Our experiments demonstrate that this novel strategy significantly boosts the performance of motion vectors, and achieves comparable performance with optical flow based action recognition methods \cite{simonyan2014two}.

The main contributions of this paper are summarized as below. Firstly, we propose a real-time CNN based action recognition method which achieves comparable performance with the state-of-the-art two-stream approach \cite{simonyan2014two}. Secondly, we firstly introduce motion vector as the input of CNN to avoid the heavy computational cost of optical flow. Finally, we propose techniques to transfer the knowledge of optical flow CNN to motion vector CNN, which significantly improves the recognition performance.


\section{Related Work}

Action recognition has been widely studied in recent years. Early approaches extracted local spatio-time descriptors from input video and encoded these descriptors with Bag of Visual Words or its variants for classification. Laptev \cite{laptev2005space} proposed spatio-time interest points by extending Harris corner into spatio-time dimension. Wang {\it et~al.} \cite{wang2013action} further exploited trajectorires to model temporal relationship of continuous frames. Furthermore, Kviatkovsky {\it et~al.} \cite{kviatkovsky2014online} proposed a covariance descriptor to realize online action recognition. Popular local descriptors for video representation include HOG \cite{dalal2005histograms}, HOF \cite{laptev2005space}, MBH \cite{dalal2006human} and TBC \cite{wang2015human}. And feature encoding techniques include hard quantization \cite{sivic2003video}, VLAD \cite{jegou2010aggregating}, and Fisher Vector \cite{sanchez2013image}. \cite{WangQT15b,WangQT13b, WangQT13a} exploited mid-level representations by proposing MoFAP and Motionlets.

Recent renaissance of deep neural network remarkably accelerates the progresses in image classification. Convolutional Neural Networks (CNNs) \cite{lecun1995convolutional} can learn powerful features from large scale image datasets, which greatly alleviates the difficulty of designing hand-crafted features. Extensive experiments have demonstrated that CNN can achieve superior performance on various image and video classification tasks, e.g. ImageNet object classification \cite{krizhevsky2012imagenet}, face recognition \cite{taigman2014deepface}, and event classification \cite{gan2015devnet}. These successes inspire researchers to extend CNN for video classification tasks \cite{karpathy2014large, simonyan2014two, tran2014c3d}. Karpathy {\it et~al.} \cite{karpathy2014large} proposed several convolutional neural network (CNN) architectures based on stacked RGB images for video classification. They designed several fusion strategy for RGB flows to utilize temporal information in stacked RGB frames. \cite{ji20133d,tran2014c3d} modeled temporal information by designing 3D filter to directly learn feature from videos. Unlike 2D filters used in image classification, 3D filters can learn temporal relationship from continuous RGB frames. Simonyan and Zisserman \cite{simonyan2014two} proposed the two-stream architecture which exploits two CNNs to model RGB and optical flow respectively. This method achieved excellent performance in practice. We will use  two-stream network as baseline of this study. Based on two-stream CNNs, Wang {\it et~al.} \cite{WangQT15a} proposed Trajectory-pooled Deep-Convolutional Descriptors to obtains the merits of CNNs and trajectory based method. Wang {\it et~al.} \cite{2015arXiv150702159W} further extended two-stream CNNs to very deep two-stream CNNs and achieve superior results on several datasets. Recent works also showed that the temporal structure in video contains discriminative information. Ng {\it et~al.}\cite{ng2015beyond} utilized the recurrent  LSTM architecture to capture temporal structure of actions. Wu {\it et~al.} \cite{wu2014exploring} showed that integrating LSTM and two-stream methods can further improve the recognition performance. The main focus of this paper is to accelerate action recognition with deep learning while preserving the high performance.


%
\section{Motion Vector for Deep Action Recognition}

Although two-stream CNNs \cite{simonyan2014two} achieve state-of-the-arts performance in action recognition, it is computational expensive and cannot be deployed for real-time process. Two-stream CNNs consist of spatial net and temporal net, which take RGB image and optical flow as input respectively. In the testing phase, feed forward computation of CNNs can be conducted in short time (around 30ms) with GPU implementation. The most computationally expensive step in the original two-stream framework comes from the calculation of optical flows. With efficient implementation \cite{kantorov2014efficient}, it takes around 360ms to calculate optical flow (Farneback's Flow \cite{farneback2003two}) for one frame in CPU. Even with  GPU acceleration \cite{simonyan2014two}, the calculation still takes 60 ms (Brox's Flow \cite{brox2011large}) which cannot meet the requirement of real-time process. Thus optical flow is the main bottleneck that prohibits the classical two-stream to be real-time.
%

Similar to optical flow, motion vector is a two-dimensional vector used for describing the moving offsets of local blocks with respect to a reference frame.  Motion vector is widely used in various video compression standards, thus can be obtained directly in video decoding process without extra calculation. These facts make motion vector an attractive feature for efficient video content analysis. Actually, the previous work \cite{kantorov2014efficient}  has used motion vectors together with VLAD encoding for action recognition and achieve good performance. Different from this work  \cite{kantorov2014efficient}, however, we explore motion vector in the deep CNN framework. Here, the major challenge comes from the fact that motion vector has low resolution and is imprecise for describing fine motions. This fact can largely harm the recognition performance of CNN if we directly train networks with motion vectors.

To relieve this problem, we propose several training methods to enhance motion vector CNN (MV-CNN) for better recognition performance. Our key insight is that the knowledge and features learned with optical flow CNN (OF-CNN) can be helpful for MV-CNN. Thus, we may leverage OF-CNN as a teacher net to guide the training of MV-CNN.




\subsection{Motion Vector}


We begin with a short introduction to motion vectors and then analyze the difficulty of training motion vector CNNs (MV-CNNs). Motion vectors are designed for describing macro blocks movement from one frame to the next, and are widely used in various video compression standards such MPEG series, HEVC. Temporal redundancy of two neighboring frames yields important cues for compressing video data.  Motion vectors exploit temporal relationship in neighboring frames by recording how macro blocks move in the next and are one of the essential ingredients in modern video coding algorithms. Figure \ref{fig:Comparison of Optical Flow and MVS} illustrates an example of motion vectors. As motion vectors are already calculated and encoded in compressed videos, we can obtain them at very low computational cost.

However, it is challenging to train MV-CNNs with high performance. This is because that motion vector is designed for video compression where precious motion information are not obligatory. Compared with optical flow, motion vectors exhibit coarse structure and may contain noisy and inaccurate movements. As shown in Figure \ref{fig:Comparison of Optical Flow and MVS}, motion vector contains macro blocks with different sizes in motion estimation, ranging from 8$\times$8 pixels to 16$\times$16 pixels. Unlike dense optical flows, which are pixel-level and provide fine movement information of single pixel, motion vectors only yield block-level motion information. Thus motion vectors exhibit much coarser structures than optical flows. Fine details contained in optical flows can not be delivered in motion vectors, such as the structure of bow in Archery videos and the baseball hold by the man's hand. These information loss can badly impact the final performance of MV-CNN.

Secondly, noisy information contained in motion vector poses a barrier for MV-CNN to achieve high performance.  Unlike modern optical flow algorithm \cite{farneback2003two,brox2011large}, motion vectors simply use three or four comparison steps to find the most matching block. There are much more noisy patterns contained in motion vectors than optical flow fields as shown in Figure \ref{fig:Comparison of Optical Flow and MVS}.
The inaccurate information and existence of noisy point are due to the fact that video compression algorithms need to balance between the speed of encoding and the compression rate.
Therefore, motion vectors can only provide noisy block movement information, which hamper performance of temporal CNN.

Thirdly, not every frame contains motion vectors. Partial reason for this is that frames are clustered as group of pictures (GOP). One typical GOP contains three types of frames: I-frame, P-frame and B-frame. I-frame is an intra-coded frame encoded based on its own, which means I-frame contains no movement information. P-frame and B-frame are acronyms of predicted frame and bi-predictive frame respectively. They contain movement information. Clearly, in action recognition with CNNs, empty I-frame can hinder CNN training process and degrade the performance. In this paper, we deal with this problem by simply replacing I-frame with the motion vectors of previous frame.

\begin{figure*}
\begin{center}
\includegraphics[width=\textwidth]{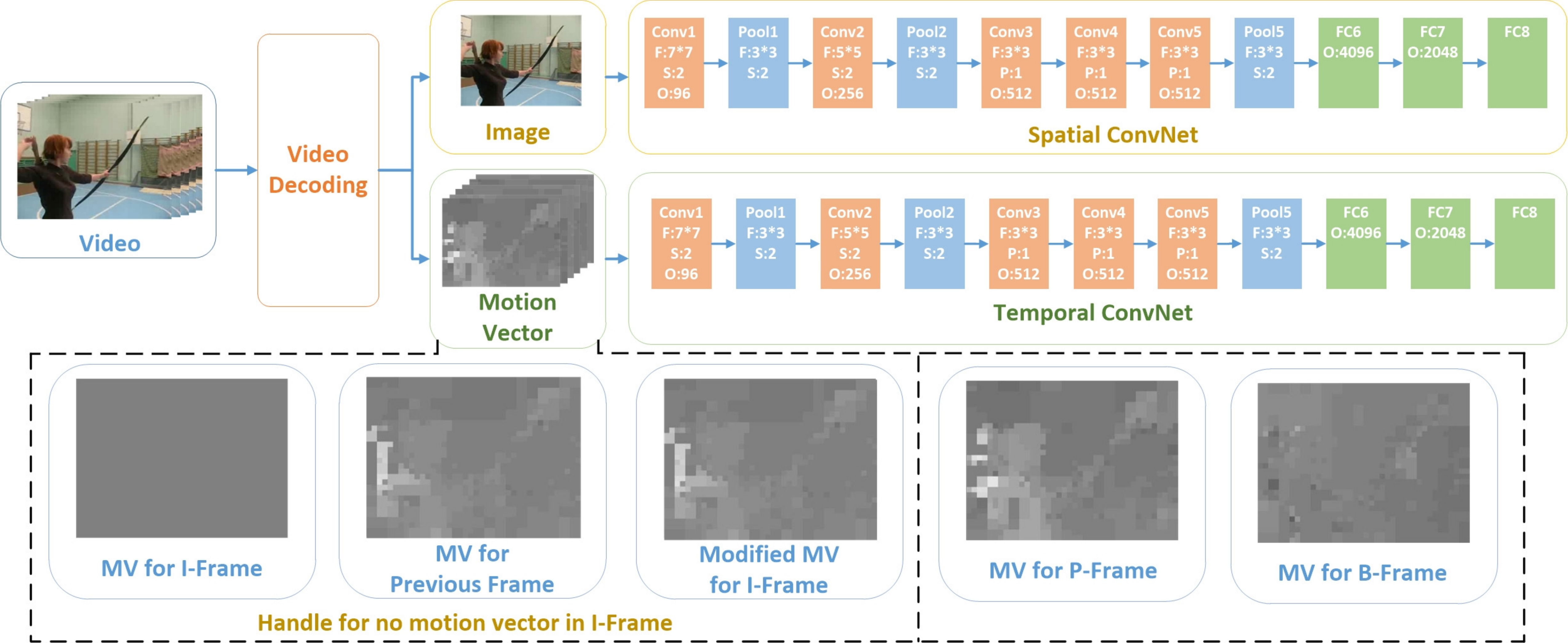}
\end{center}
   \caption{Structure for real-time action recognition system. In spatial and temporal CNN, F stands for kernel size and S means stride step. O represents for output number and P is pad size.
   }

\label{fig:Real-time Action Recognition Structure}
\vspace {-2mm}
\end{figure*}

\subsection{Real-time Action Recognition Frameworks}

Our proposed real-time action recognition framework contains two components (Figure \ref*{fig:Real-time Action Recognition Structure}). The first component is video decoder, which extracts RGB images and motion vectors from input compressed videos. The second component follows the two-stream architecture \cite{simonyan2014two}, which can be decomposed into spatial CNN (RGB-CNN) and temporal CNN (MV-CNN). The main difference comes from the fact that we use motion vector as input for temporal CNN while two-stream uses optical flows. As both RGB images and motion vectors can be obtained directly from the video decoding process, our method avoids the computationally expensive step to estimate optical flow, which is most time-costly in the original two-stream framework.

In the training phase, we extract RGB images and frames of motion vectors from video. These images and motion vectors inherit the labels of original videos. To augment training samples, we crop image and MV frames in spatial domain. Then we train RGB-CNN and MV-CNN with these cropped samples, respectively. In the testing phase, raw images and motion vector frames from testing video are fed forward into RGB-CNN and MV-CNN. The action recognition decision will be made by weighted average of prediction scores from two CNNs. The weight for spatial and temporal CNN are set as 1 and 2.

We use the architecture of ClarifaiNet \cite{zeiler2014visualizing} for both RGB-CNN and MV-CNN, since ClarifaiNet keeps a balance between efficiency and performance. For spatial net, dropout layers are added after FC6 and FC7 layer. We follow \cite{simonyan2014two} by setting dropout ratio for spatial CNN to 0.9 to avoid over-fitting. Spatial net is pre-trained on ImageNet ILSVRC-2012 datasets. Batch size for spatial and temporal nets are set as 256. The learning rate is initialized as  $10^{-3}$ and decreases into $10^{-4}$ after 14k steps. The training procedure stops at 20k iterations.

Our temporal CNN is slightly different with ClarifaiNet by replacing all ReLU layers with PReLU \cite{KaimingHE} layers, which exhibits better performance and quick convergence. Dropout ratio for temporal net is set to 0.9 and 0.8 after FC6 and FC7 layer respectively. Temporal CNN is trained from scratch by stacking 10-frame motion vectors as input. Its learning rate starts from $10^{-2}$, and drops to $10^{-3}$ at 30k steps. We then decrease the learning rate to $10^{-4}$ after 70k iterations. The whole learning process stops at 90k steps.

\section{Enhanced Motion Vector CNNs}

\begin{figure*}
\begin{center}
\subfigure[Strategy 1: Teacher Initialization]{\includegraphics[width=0.24\textwidth]{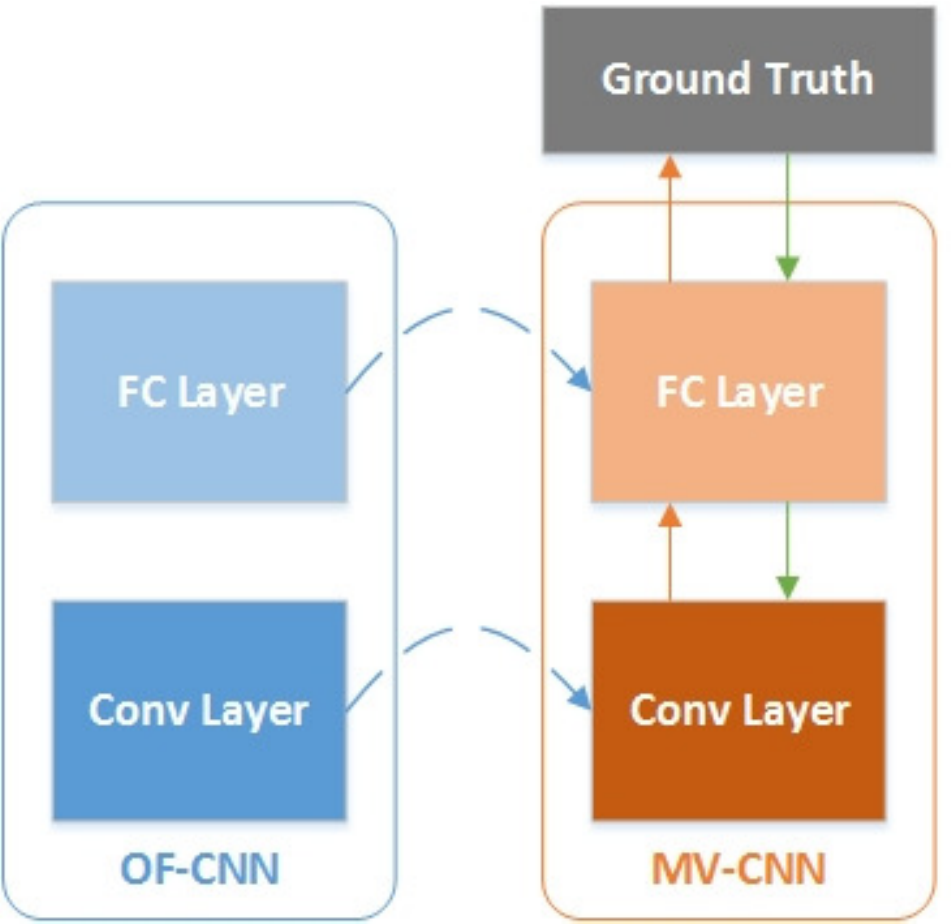}}
\subfigure[Strategy 2: Supervision Transfer]{\includegraphics[width=0.3\textwidth]{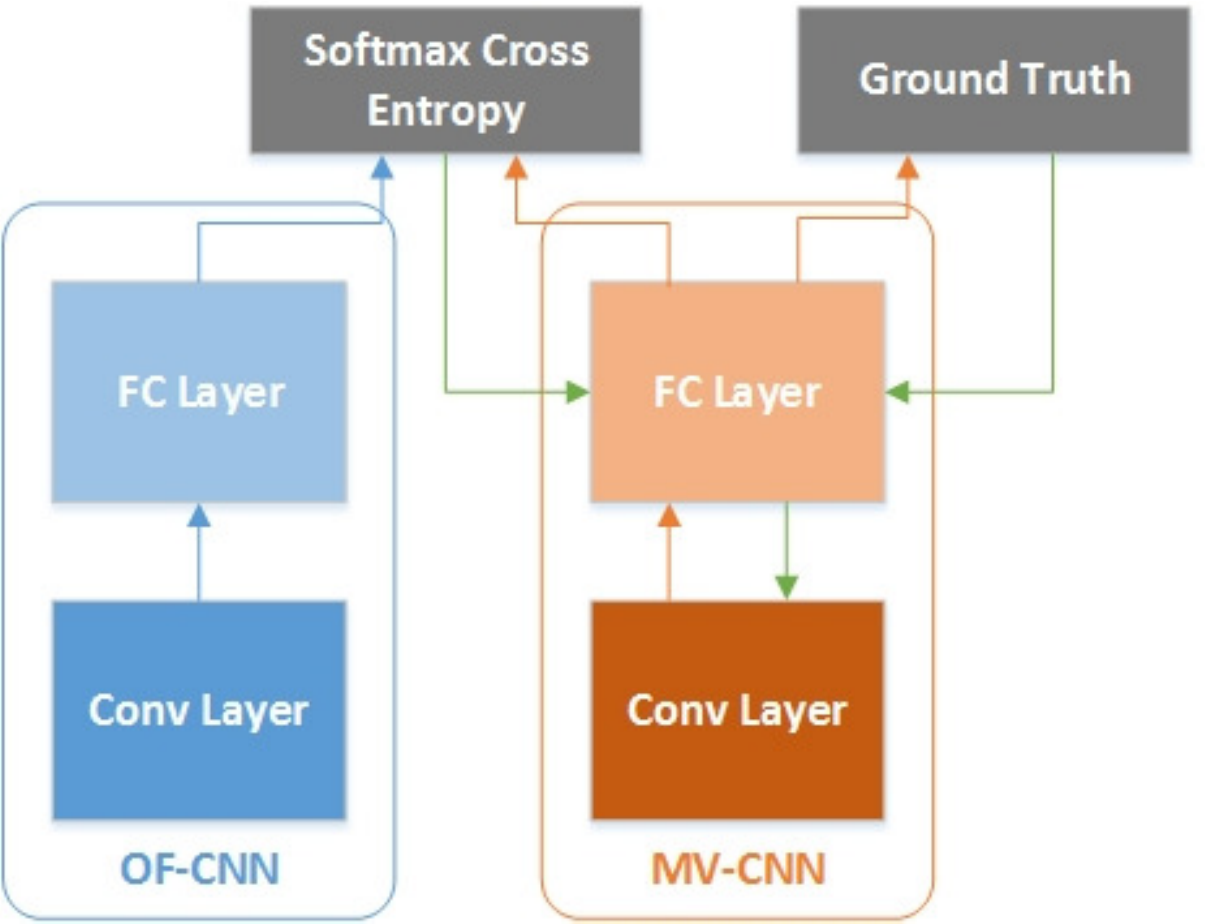}}
\subfigure[Strategy 3: Combination]{\includegraphics[width=0.3\textwidth]{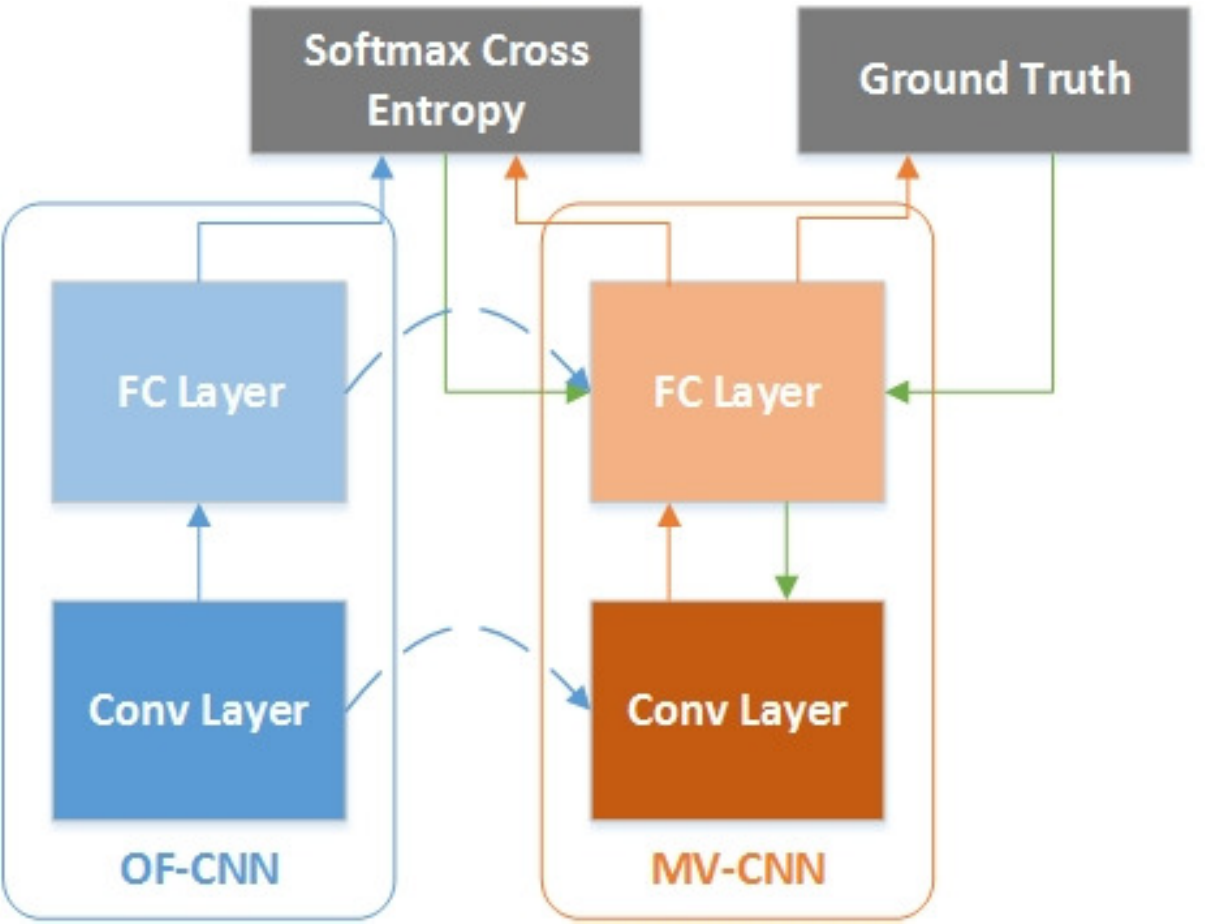}}
\end{center}
\vspace {-2 mm}
\caption{Structure of three strategies. Blue dash lines represent copying the initial weights from teacher net to student net. Green lines are the backward propagation path. Blue full lines mean feed forward paths of teacher flow. Orange lines are feed forward paths of student net.}
\label{fig:Polishing CNN}
\vspace {-2 mm}
\end{figure*}

As analyzed above, motion vectors lack fine details and contain noisy and inaccurate motion patterns, which makes training motion vector CNN (MV-CNN) more challenging. We observe that simply replacing optical flow with motion vector can lead to significant recognition performance degradation of around 7\%. In this section we try to keep the high speed merit of motion vectors while achieve the high performance as optical flow. Here our key insight to circumvent this difficulty is that  motion vector and optical flow are inherent correlated to each other. This fact enables us to leverage the rich knowledge and fine features learned in optical flow CNN (OF-CNN) to enhance MV-CNN. This can be seen as a knowledge transfer problem from optical flow domain to motion vector domain. More specifically, we take OF-CNN as a teacher net and use it to teach the MV-CNN net. It should be noticed that OF-CNN teaches MV-CNN only in the training phase. Thus we do not need to calculate optical flows in testing phase, and the system speed performance will not be affected by the proposed algorithm.

In particular, we propose three strategies to transfer knowledge from OF-CNN to MV-CNN. To begin with, several notations are introduced at first.
Parameters for teacher CNN in optical flow domain is denoted by $T_{p}=\{T_{p}^1,T_{p}^2,...,T_{p}^n\}$, where $n$ represents the total number of layers. As for student CNN (MV-CNN), its parameter is defined as $S_{p}=\{S_{p}^1,S_{p}^2,...,S_{p}^n\}$. In this paper, we assume MV-CNN has the
same network structure as OF-CNN, while the techniques can be easily generalized to those with different structures.

\subsection{Teacher Initialization}

Extensive works show that the initialization of network parameters can largely impact the final performance.
Both optical flow and motion vector describe the motion information of local regions, and are inherently correlated. This fact inspires us to initialize the parameters of MV-CNN as those of its teacher's net OF-CNN,
\begin{equation}
S_{p}^t = T_{p}^t, t=1,...,n.
\end{equation}

Then we fine-tune the parameters of MV-CNN  with the motion vector samples until convergence. This process can also be seen as pre-training MV-CNN with the fine optical flow samples, which transfers the knowledge learned by teacher net to student net directly.

For implementation details, the learning rate is initiated from $10^{-3}$, and then drops to $10^{-4}$ and $10^{-5}$ at 30k and 70k steps respectively. The training stops at 90k iterations.


\subsection{Supervision Transfer}

Initialization of MV-CNN with OF-CNN's parameters is simple, but the initial knowledge transferred to MV-CNN may be lost during the fine-tuning process with motion vector samples. To relieve this problem, we introduce the supervision transferring approach which takes account of additional supervision from the teacher net in the training process of MV-CNN. OF-CNN can extract effective representation from input frame. Here we take the representation obtained in the FC layer of OF-CNN as a new supervision for training MV-CNN.

This technique is in spirit similar to Hinton {\it{et~al.}}'s  recent work on knowledge distillation \cite{hinton2015distilling}.  The aim of knowledge distillation is to compress a cumbersome network (teacher) into a small network (student) which achieves similar performance as the cumbersome one. The cumbersome network and small network have the same input. Different from this work, in our problems, the teacher (OF-CNN) and the student (MV-CNN) take different types of input, {\it i.e.} optical flow vs. motion vector, but have the same network structure. Moreover, our objective is not to compress the student to a small net, but to enhance the performance of the student with low quality input.

Formally, for a given frame $I$ with optical flow feature $o$ and motion vector $v$ , we calculate the output of the last FC layer of teacher CNN as: $T^n(o)=\text{softmax}(T^{n-1}(o)) ,$
where `softmax' function is used to transform the feature $T^{n-1}$ to a probability score of multiple classes.
Similarly, the output of student's last layer is defined as: $S^n(v)=\text{softmax}(S^{n-1}(v))$.

To transfer knowledge from teacher to student, we hope $S^n(v)$ can approximate $T^n(o)$ as closely as possible. We introduce a $teacher~supervision~loss$ function to minimize the difference between $S^n(v)$ and $T^n(o)$. The cross-entropy loss is used to measure the difference. Following \cite{hinton2015distilling}, we introduce a temperature $Temp$ to soften the next-to-last layer output. The softmax output of teacher net is softened as $P_T=\text{softmax}(T^{n-1}/Temp) ,$
Student net's softmax output for second target is similarly defined as:
$P_S=\text{softmax}(S^{n-1}/Temp) ,$
Then the teacher supervision loss (TSL) with cross-entropy function is defined by :
\begin{equation}
L_{TSL}=-\sum_{i=1}^{k} P_T(i) \log P_S(i),
\label{TSL}
\end{equation}
where $k$ is the dimension of the student's output (same as that of the teacher).

In addition to the teacher supervision loss, we also minimize the cross entropy between student's output and the ground truth $Q$, which is given by,
\begin{equation}
L_{GT}=-\sum_{i}\mathbbm{1}[Q=i]\log S^n(i) ,
\label{GTL}
\end{equation}
where $S^n$ and $Q$ represent the hard output distribution vectors and the ground truth label respectively.

Our final loss function combines the teacher supervision loss (Eq.\ref{TSL}) and the ground truth loss (Eq.\ref{GTL}):
\begin{equation}
L=L_{TSL}+ w \times L_{GT}
\label{TL}
\end{equation}
where $w$ is a weight to balance these two terms. Thus, the student MV-CNN can receive supervision signal from both the teacher OF-CNN and the ground truth. It should be noted that in supervision transfer, the weights of teacher model are frozen. In this way, the knowledge is transferred from the teacher net to the  student one.

For implementation details,  we set the initial learning rate as $10^{-3}$. We decay the learning rate to $10^{-4}$ at 50k and $10^{-5}$ at 70k steps. The whole training procedure stops at 90k iterations.

\subsection{Combination}

In the third strategy, we combine the initialization with teacher's parameters and supervision transfer to further enhance the performance of student's net. We first train the teacher OF-CNN with optical flow fields. Then we initialize the student MV-CNN  with OF-CNN's parameters. After this, we train the student net with supervision signal from both the teacher and the ground truth (Eq. \ref{TL}). In this way, the student net not only inherits teacher's parameters from initialization, but also mimics teacher net's prediction during fine tuning procedure. This allows us to combine the merits of two previous strategies and further boost the generalization ability of student MV-CNN.


\begin{table}
\begin{center}
\begin{tabular}{l|c}
\hline
Temporal CNN & Accuracy \\
\hline
\hline
OF-CNN \cite{simonyan2014two} & 81.2\% \\
\hline
MV-CNN trained from scratch & 74.4\% \\
EMV-CNN with ST & 77.5\% \\
EMV-CNN with TI & 78.2\% \\
EMV-CNN with ST+TI & \bf{79.3\%} \\
\hline
\end{tabular}
\end{center}
\caption{Comparison of temporal CNN performance for Optical Flow based approach and Motion vector based Method on UCF101 (Split1). ST stands for Supervision Transfer and TI means Teacher Initialization.}
\label{UCF101 split1 performance Temporal Net}
\end{table}

\section{Experiment}

In this section we first describe the evaluation datasets. Then we report and analyze the experimental results.

\subsection{Datasets and Evaluation Protocol}

The evaluation is conducted on two challenging datasets: UCF101~\cite{soomro2012ucf101} and THUMOS14~\cite{THUMOS14}. UCF101 contains 13,320 videos. The datasets has three splits for training and testing. We follow the standard setup and report average accuracy over three splits on this datasets.

THUMOS14 is a dataset for action recognition challenge 2014. It contains 13,320 videos for training, 1,010 videos for validation, and 1,574 videos for testing. Unlike UCF101, THUMOS14 uses untrimmed videos for validation and testing. Lots of irrelevant frames make the training and testing of CNNs more difficult. We use training and validation datasets to train our CNNs. Official evaluation tool is utilized to evaluate our system performance. According to the standard setup of this dataset, we report the mean Average Precision (mAP) on testing dataset. As videos in THUMOS14 are untrimmed and have large number of frames, we conduct CNN testing at every 20 frames.

For experiments on both datasets, the speed evaluation is measured as frames per seconds (fps) on a single-core CPU (E5-2640 v3) and a K40 GPU.

\subsection{Implementation Details}

In order to learn robust features from CNNs, we use three data augmentation strategies. Firstly, we randomly crop a $224 \times 224$ patch from image set. Random cropping can provide more training data for CNNs to learn better features. Secondly, we horizontally flip cropped patches by random. Furthermore, following \cite{2015arXiv150702159W}\footnote{\url{https://github.com/yjxiong/caffe}}, we use a scale jittering strategy to help CNN to learn robust features. We crop a patch from dataset on three scales 1, 0.875, and 0.75, which yield patches of size $224 \times 224$, $196 \times 196$ and $168 \times 168$ respectively. The patches are then resized to $224 \times 224$ after this multi-scale strategy. In testing phase, we crop one $224 \times 224$ patch from the center of testing image. No data augmentation strategy is used in evaluation phase.

Our teacher CNN is trained on TV-L1 optical flow \cite{TVL1} that achieves 81.6\% on UCF101 Split1, which is comparable with the performance in the original paper 81.2\% \cite{simonyan2014two}.

\begin{table}
\begin{center}
\begin{tabular}{l|c}
\hline
CNN & MAP \\
\hline
\hline
RGB CNN & 57.7\% \\
OF-CNN & 55.3\% \\
RGB CNN+OF-CNN & 66.1\% \\
\hline
MV-CNN & 29.8\% \\
EMV-CNN & 41.6\% \\
\hline
RGB CNN+MV-CNN & 58.7\%\\
RGB CNN+EMV-CNN & \bf{61.5\%}\\
\hline
\end{tabular}
\end{center}
\caption{Performance of EMV-CNNs and MV-CNNs on THUMOS 14 dataset. We also report the results of two-stream CNNs.}
\label{TH14 performance Temporal Net}
\end{table}

\subsection{Parameter Sensitivity}

\begin{table}
\begin{center}
\begin{tabular}{l|c|c|c}
\hline
 & MV & CNN & Total \\
Dataset & (fps) & (fps) & (fps) \\
\hline
\hline
UCF101 & 735.3 & 833.3 & 390.7 \\
THUMOS14 & 781.3 & 833.3 & 403.2 \\
\hline
\end{tabular}
\end{center}
\caption{Speed of each components in Real-time Action Recognition System. MV stands for motion vector extraction, while CNN means convolutional neural network processing.}
\label{Speed of each components in Real-time Action Recognition System}
\end{table}

\begin{table}
\begin{center}
\begin{tabular}{l|c|c|c}
\hline
 &  Spatial  & Brox's Flow\cite{brox2011large} & MV \\
Dataset & Resolution & (GPU) (fps) & (CPU) (fps) \\
\hline
\hline
UCF101 & $320 \times 240$ & 16.7 & 735.3 \\
THUMOS14 & $320 \times 180$ & 17.5 & 781.3 \\
\hline
\end{tabular}
\end{center}
\caption{Comparison of speed for optical flow fields and motion vectors. MV means motion vector.}
\label{Comparison of speed for optical flow fields and motion vectors}
\end{table}

We first analyze the parameter sensitivity. There are two important parameters in our real-time action recognition system: temperature $Temp$ for Supervision Transfer and weight $w$ for soft target. Following \cite{hinton2015distilling} we set soft target weight $w=Temp^2$ to balance the gradients of soft target. We test three different temperature settings on UCF101 split1 by setting $Temp$ to 1, 2 and 3. The corresponding soft target weight $w$ is set as 1, 4 and 9. Accuracy grows up from $79.2\%$ to $79.3\%$ from temperature 1 to 2 and goes down to $78.9\%$ at temperature 3. We can see that the recognition results of different temperatures are very close, which implies that our method is robust to temperature. According to this study, we set $Temp=2$ in the following experiments.

\subsection{Evaluation of MV-CNNs and EMV-CNNs}

In this subsection, we compare and analyze different training strategies for motion vector CNN (MV-CNN) on UCF101 Split1 and THUMOS14. The result are shown in Table \ref{UCF101 split1 performance Temporal Net} and Table \ref{TH14 performance Temporal Net}. As \cite{simonyan2014two} did not provide results on THUMOS14, we re-implement two-stream CNNs on this dataset.

First, from these results, we can conclude that directly using motion vector to replace optical flow  degrades system's performance. Compared with OF-CNN, MV-CNN trained from scratch degrades performance by 7\% and 25\%, which achieves 74.4\% and 29.8\% on UCF101 Split1 and THUMOS14 respectively. It indicates that coarse structure and imprecise movement information in motion vector harm the performance of CNNs. The performance on THUMOS14 is particularly large, as validation and testing videos in THUMOS14 are untrimmed and labels for a large part of frames are not correct. Furthermore, lots of shots shift in videos aggravate the difficulties of training MV-CNN on THUMOS14 dataset.

Second, comparing Enhanced Motion Vector CNN (EMV-CNN) and MV-CNN, we observe a performance improvement of 4.9\% and 12\% on UCF101 Split1 and THUMOS14 respectively. This indicates that our newly designed method for training EMV-CNN is effective and can improve the generalization ability of MV-CNNs.

Finally, combined with spatial CNN, EMV-CNN still outperforms MV-based CNN, which indicates that the knowledge of EMV-CNN can be more complementary to spatial CNN. In addition, combining EMV-based with spatial CNN exhibits only a minor performance loss compared with OF-based CNN on UCF101. Specifically, we study the effect of our proposed transferring techniques as follows.

Teacher initialization technique can provide an improvement of 3.8\%. From this result, we can see, similar to the study in image classification \cite{krizhevsky2012imagenet, simonyan2014two}, teacher initialization on OF-CNN can benefit the MV-CNN by providing a good initial point to train. Supervision transfer strategy can provide 3.1\% performance improvement on UCF101 Split1. It shows that knowledge provided by OF-CNN during supervision transfer process in training phase can be helpful to MV-CNN. Combing supervision transfer and teacher initialization strategy can further boost the performance. As indicated in \cite{hinton2015distilling}, supervision transfer can be used to regularize MV-CNN from over-fitting. Supervision strategy utilizes the soft codes produced by OF-CNNs during training process. These codes are further softened by dividing $Temp$. Unlike hard ground truth labels, these soft targets encode more rich information and guide the training of EMV-CNN, contributing to improve the generalization ability of EMV-CNNs.

\subsection{Speed Evaluation}

\begin{table}
\begin{center}
\begin{tabular}{c|c|c}
\hline
 & UCF101(Split1) & THUMOS14 \\
EMV+RGB-CNN & (fps) & (fps) \\
\hline
\hline
5 crops+mirror & 88.0 & 89.0 \\
1 crop & 390.7 & 403.2 \\
\hline
\end{tabular}
\end{center}
\caption{Speed comparison for 1 crop and 5 crops+mirror in UCF101 and THUMOS14. EMV+RGB-CNN stands for two-stream based CNN with EMV-CNN and RGB-CNN.}
\label{Speed comparison of 1 crop and 10 crops in UCF101 and THUMOS14}
\end{table}

\begin{table}
\begin{center}
\begin{tabular}{c|c|c}
\hline
EMV+RGB-CNN & UCF101(Split1) & THUMOS14 \\
\hline
\hline
5 crops+mirror & 86.6\% & 61.2\% \\
1 crop & 85.7\% & 61.5\% \\
\hline
\end{tabular}
\end{center}
\caption{Comparison of performance for 1 crop and 5 crops+mirror in UCF101 and THUMOS14. EMV+RGB-CNN stands for two-stream based CNN with EMV-CNN and RGB-CNN.}
\label{Comparison of 1 crop and 10 crops in UCF101 and THUMOS14}
\end{table}

\begin{table}
\begin{center}
\begin{tabular}{l|c|c}
\hline
 & Accuracy & FPS\\
\hline
\hline
MV+FV (CPU) (re-implement) \cite{kantorov2014efficient} & 78.5\% & 132.8 \\
C3D (1 net) (GPU) \cite{tran2014c3d} & 82.3\% & 313.9 \\
C3D (3 net) (GPU) \cite{tran2014c3d} & 85.2\% & - \\
iDT+FV (CPU) \cite{wang2013action} & 85.9\% & 2.1 \\
Two-stream CNNs (GPU) \cite{simonyan2014two} & 88.0\% &  14.3 \\
\hline
EMV+RGB-CNN & {\bf 86.4\%} & \bf 390.7 \\
\hline
\end{tabular}
\end{center}
\caption{Comparison of speed and performance with state-of-the-art on UCF101}
\label{Speed and result comparison on UCF101}
\vspace {-1mm}
\end{table}
In this subsection, we analyze the speed of different components in our action recognition approach. In our implementation, We use CPU to extract motion vector while use GPU to conduct the feed forward process of CNN. It should be noticed that our system process a volume of 10 frames of MV and 1 frame of RGB together. Our speed is measured based on frames instead of volume.

We first compare the speed performance of 1 crop with 5 crops+mirror on UCF101 and THUMOS14 in Table \ref{Speed comparison of 1 crop and 10 crops in UCF101 and THUMOS14} and Table \ref{Comparison of 1 crop and 10 crops in UCF101 and THUMOS14}. We can observe that 1 crop only achieves a slight performance degradation than 5 crops+mirror, while 1 crop is 4 times faster than the latter.

Next, we evaluate speed performance of each component on UCF101 datasets and THUMOS14 datasets. The spatial resolution of video in UCF101 dataset is $320 \times 240$, while for video in THUMOS14, its resolution is $320 \times 180$. As shown in Table  \ref{Speed of each components in Real-time Action Recognition System}, our action recognition system achieves 390.7 fps and 403.2 fps on UCF101 and THUMOS14 respectively, which is one order faster than real-time requirement (25 fps).

Finally, we compare the efficiency of extracting motion vectors and optical flow in Table \ref{Comparison of speed for optical flow fields and motion vectors}. Motion vector extraction is almost 30 times faster than real-time requirement. Although the spatial resolutions for videos in UCF101 and THUMOS14 are different, the time consumption for extraction of motion vectors are similar. The estimation of motion vector in CPU is 44 times faster than calculating Brox's flow\cite{brox2011large} with GPU. The calculation of optical flows poses the main bottleneck in boost the speed of classical two-stream framework, which prohibits it to be conducted in real-time.

\begin{figure}[t]
\begin{center}
   \includegraphics[width=0.9\linewidth]{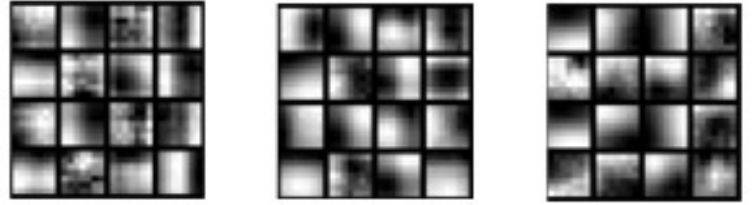}
\end{center}
\vspace{-1mm}
   \caption{Samples of filters for Conv1 layer. Left to right: MV-CNN, EMV-CNN and OF-CNN.}
\label{fig:filter}
\vspace{-2mm}
\end{figure}

\subsection{Comparison with the State of the Art}

In this section, we compare our method with several state-of-the-art methods. Since the I/O is related to hardware and operating system, we report computational time cost without I/O. Unlike those using Support Vector Machine (SVM) to perform classification \cite{tran2014c3d,wang2013action}, our action recognition approach has an {\it end-to-end structure} that combines feature extraction and classification together.

We first compare speed and accuracy performance on UCF101 (3 Splits) dataset. Results are given in Table \ref{Speed and result comparison on UCF101}. Our method is 3 times and 180 times faster than motion vector + fisher vector (MV FV) and iDT+FV respectively. It should be noted that this may not be a fair comparison since iDT and FV are implemented in CPU. At the same time, our method achieves higher accuracy than these methods.
For MV FV, we use the public code offered by \cite{kantorov2014efficient}. Although iDT used Farneback optical flows \cite{farneback2003two} which provides more precise movement information than motion vector, our methods still obtain higher performance than iDT.

We also make comparison with deep learning methods, namely two-stream CNNs and C3D.  Our method achieves {\bf 390.7} fps on UCF101 and is the fastest among all methods compared (Table \ref{Speed and result comparison on UCF101}). And our recognition accuracy is 4.1\% and 1.2\% higher than C3D (1 net) and C3D (3 net).
As for two-stream CNNs, our method achieves comparable results but ours is 27 times faster than them.

Finally, we compare our speed and mean average precision with \cite{jain201515} on THUMOS14 dataset (Table \ref{Speed and result comparison on THUMOS14}). Our method achieves better performance than Objects for 16.8\% and obtains comparable performance with Motion (iDT) and iDT+CNN, but exhibits worse performance than Objects+Motion. However, as Objects+Motion is built on the iDT features, our method is 200 times faster than it.

\begin{table}
\begin{center}
\begin{tabular}{l|c|c}
\hline
 & Accuracy & FPS\\
\hline
\hline
Objects (GPU) \cite{jain201515}& 44.7\% & - \\
iDT+CNN (CPU+GPU) \cite{WangQT14} & 62.0\% & $ < 2.38$ \\
Motion (iDT+FV) (CPU) \cite{jain201515}& 63.1\% &  2.38\\
Objects+Motion (CPU+GPU) \cite{jain201515}& 71.6\% &$ < 2.38$ \\
\hline
EMV+RGB-CNN & {\bf 61.5\%} & {\bf 403.2} \\
\hline
\end{tabular}
\end{center}
\caption{Comparison of speed and performance with state-of-the-art on THUMOS14}
\label{Speed and result comparison on THUMOS14}
\vspace{-3mm}
\end{table}

\subsection{Visualization of Filters}

In order to further explore the effectiveness of EMV-CNN, we visualize filters of the first layer (Conv1) for MV-CNN, EMV-CNN and OF-CNN in Figure \ref {fig:filter}.
It can be clearly noticed that filters of MV-CNN contain coarse and noisy information, which may be ascribed to the fact that  motion vector lacks fine grained information as shown in Figure \ref {fig:Comparison of Optical Flow and MVS}. Compared with filters of MV-CNN, EMV-CNN can obtain detailed information, which shows that our proposed learning scheme successfully transfer knowledge from OF-CNN to EMV-CNN. Filters of EMV-CNN are cleaner than MV-CNN and have fine structures, contributing to improve recognition performance.

\section{Conclusions}

In this paper we have proposed a motion vector CNN to accelerate the speed of deep learning methods for action recognition. Motion vectors can be extracted directly in video decoding process without extra computation. However, motion vectors lacks fine and accurate motion information which degrades recognition performance. To relieve this problem, we developed three knowledge transfer techniques to adapt the models of optical flow CNN to motion vector CNN, which significantly boost the recognition performance of the latter. Our method achieves 391 fps and 403 fps with high performance on UCF101 and THUMOS14 respectively.

\section {Acknowledgement}

This work is partly supported by National Natural Science
Foundation of China (61472410, 61472281), Guangdong Innovative Research Program (2015B010129013, 2014B050505017), Shenzhen Research Program (KQCX2015033117354153,  JSGG20150925164740726, CXZZ20150930104115529) and the Program for Professor of Special Appointment (Eastern Scholar) at the Shanghai Institutions of Higher Learning (GZ2015005).

{\small
\bibliographystyle{ieee}
\bibliography{egbib}
}

\end{document}